\documentclass{article}

\usepackage[utf8]{inputenc} 
\usepackage[T1]{fontenc}    
\usepackage{hyperref}       
\usepackage{url}            
\usepackage{booktabs}       
\usepackage{amsfonts}       
\usepackage{nicefrac}       
\usepackage{microtype}      
\usepackage{xcolor}         

\usepackage{caption}
\usepackage{subcaption}
\usepackage{amsmath} 
\usepackage{amssymb}
\usepackage{subfiles}
\usepackage{breqn}
\usepackage{amsmath}
\usepackage{mathtools}
\usepackage{algorithm,algpseudocode}
\usepackage{wrapfig}
\usepackage{natbib}
\usepackage{multirow}
\newcommand{\AB}{\mathbf{A}}
\newcommand{\BB}{\mathbf{B}}

\newcommand{\DB}{\mathbf{D}}

\newcommand{\GB}{\mathbf{G}}

\newcommand{\IB}{\mathbf{I}}

\newcommand{\KB}{\mathbf{K}}
\newcommand{\LB}{\mathbf{L}}

\newcommand{\SB}{\mathbf{S}}
\newcommand{\TB}{\mathbf{T}}

\newcommand{\WB}{\mathbf{W}}

\newcommand{\YB}{\mathbf{Y}}

\newcommand{\oneB}{\mathbf{1}}

\newcommand{\pB}{\mathbf{p}}
\newcommand{\qB}{\mathbf{q}}

\newcommand{\uB}{\mathbf{u}}
\newcommand{\vB}{\mathbf{v}}

\newcommand{\xB}{\mathbf{x}}
\newcommand{\yB}{\mathbf{y}}
\newcommand{\zB}{\mathbf{z}}

\newcommand{\RBB}{\mathbb{R}}

\newcommand{\AM}{\mathcal{A}}

\newcommand{\CM}{\mathcal{C}}
\newcommand{\EM}{\mathcal{E}}

\newcommand{\GM}{\mathcal{G}}

\newcommand{\NM}{\mathcal{N}}
\newcommand{\OM}{\mathcal{O}}

\newcommand{\VM}{\mathcal{V}}

\newcommand{\XM}{\mathcal{X}}
\newcommand{\YM}{\mathcal{Y}}

\newcommand{\PsiB}{\mbox{\boldmath$\Psi$\unboldmath}}

\newcommand{\muB}{\mbox{\boldmath$\mu$\unboldmath}}

\newcommand{\DeltaB}{\mbox{\boldmath$\Delta$\unboldmath}}

\newcommand{\argmax}{\mathop{\rm argmax}}
\newcommand{\argmin}{\mathop{\rm argmin}}

\newcommand{\trace}{\operatorname{trace}}
\newcommand{\proj}{\operatorname{Proj}}
\newcommand{\KL}{\operatorname{KL}}
\newcommand{\Diag}{\operatorname{Diag}}

\newcommand{\dataset}{dataset}
\newcommand{\datasets}{datasets}
\newcommand{\groundtruth}{ground truth}

\newcommand{\realworld}{real-world}

\newcommand{\Datasets}{Datasets}

\newcommand{\Realworld}{Real-world}

\title{\method{}: A Structural Inconsistency Reducing Graph Matching Algorithm}

\author {
	
	Weijie Liu,\textsuperscript{\rm 1}
	Chao Zhang, \textsuperscript{\rm 1}
	Nenggan Zheng, \textsuperscript{\rm 1}
	Hui Qian \textsuperscript{\rm 1}\\
}
\date{%
	\textsuperscript{\rm 1} Zhejiang University \\%
	\{westonhunter, zczju, zng, qianhui\}@zju.edu.cn\\[2ex]%
	\today
}

\newtheorem{theorem}{Theorem}
\newtheorem{proposition}[theorem]{Proposition}
\newtheorem{lemma}[theorem]{Lemma}

\newcommand{\src}{\operatorname{s}}
\newcommand{\tar}{\operatorname{t}}

\newcommand{\MNC}{\operatorname{MNC}}

\newcommand{\st}{\text{ s.t. }}

\newcommand{\maNC}{\operatorname{NC}}
\newcommand{\mareal}{\operatorname{real}}

\newcommand{\SI}{Structural Inconsistency}
\newcommand{\si}{structural inconsistency}
\newcommand{\sis}{SI}

\newcommand{\masis}{\operatorname{SI}}
\newcommand{\macsi}{\operatorname{CSI}}
\newcommand{\method}{SIGMA}

\begin{document}

\maketitle


\begin{abstract}
	Graph matching finds the correspondence of nodes across two correlated graphs and lies at the core of many applications.
	When graph side information is not available,
	the node correspondence is estimated on the sole basis of network topologies.
	In this paper, we propose a novel criterion to measure the graph matching accuracy, \emph{\si{}} (\sis{}), which is defined based on the network topological structure.
	Specifically, \sis{} incorporates the heat diffusion wavelet to  accommodate the multi-hop structure of the graphs.
	Based on \sis{}, we propose a \emph{Structural Inconsistency reducing Graph Matching Algorithm} (SIGMA),
	which improves the alignment scores of node pairs that have low \sis{} values in each iteration.
	Under suitable assumptions, \method{} can reduce \sis{} values of true counterparts.
	Furthermore, we demonstrate that \method{} can be derived by using a mirror descent method to solve the Gromov-Wasserstein distance with a novel $K$-hop-structure-based  matching costs.
	Extensive experiments show that our method outperforms state-of-the-art methods.
\end{abstract}

\section{Introduction}
Graph matching (network alignment) aims to identify the correspondence of nodes across two related graphs,
by minimizing the sum of pair-wise matching costs of corresponding nodes/edges.
It has many applications,
such as
linking user accounts in social platforms \citep{shu2017user,wang2019user},
aligning entities in knowledge graphs \citep{sun2020knowledge,pei2020rea},
and matching keypoints across two images \citep{sarlin2020superglue,wang2020zero},
to name a few.


The choice of matching costs is critical to the performance of graph matching.
When  node/edge side information, e.g. attributes, is available,
one can rely on expert knowledge to design handcrafted node embeddings for the construction of matching costs
\citep{zhang2016final,meng2016igloo}.
Recently, to alleviate the requirement of expert knowledge, 
end-to-end deep learning methods are proposed
to learn the node embedding automatically based on the \emph{anchor links} (labelled node correspondences) \citep{zanfir2018deep,fey2019deep,wang2019learning,sarlin2020superglue,wang2020combinatorial}.
However, in many tasks, side information may not be available\citep{snapnets,heimann2021refining}.
Under such circumstance,
the matching costs can merely be estimated on the basis of the network topologies of the two graphs \citep{patro2012global,saraph2014magna,sun2015simultaneous,memoli2011gromov,maretic2019got,xu2019gromov,heimann2021refining}.

Among these network-topology based methods,
one line of works align graphs by borrowing tools from optimal transport (OT) and achieve the state-of-the-art performance  \citep{memoli2011gromov,maretic2019got,xu2019gromov,titouan2019optimal,chowdhury2019gromov,barbe2020graph}.
Basically, these methods utilize OT to exploit the geometric properties of the metric spaces that underlie the graphs, 
and  estimate the node correspondence by calculating the \emph{transport plan} of the Gromov-Wasserstein distance. 
Typically, the matching costs in these methods are calculated depending on the one-hop neighborhood information of each node.

Recently, \citet{heimann2021refining} reveals the connection between the graph matching accuracy and the  \emph{matched neighborhood consistency} (MNC), i.e., the Jaccard simliarity between the one-hop neighborhoods of matched node pairs.
In particular, they show that accurate matching essentially entails high MNC.
However, it remains unknown how the OT-based methods  are related to MNC theoretically.
Instead, they propose a method, RefiNA, by maximizing the numerators of MNC, which can be regarded as an opposite measure of the matching cost.
Though the authors intend to find matching with high MNC, RefiNA actually diverges from their purpose, as the denominator of MNC is ignored and a high numerator does not ensure a high MNC.
Moreover, both OT-based methods and RefiNA may suffer from one-hop structural indistinguishability and tend to misalign nodes that have low degrees, as they discard the abundant network topological information and estimate the graph matching merely depending on the one-hop neighborhoods structure.

In this paper, we propose a novel criterion to measure the graph matching accuracy, \emph{\si{}} (\sis{}), which utilizes the abundant network topological information.
Specifically, \sis{} incorporates the \emph{heat diffusion wavelet} \citep{donnat2018learning} to  accommodate the multi-hop structure of the graphs.
If a pair of nodes incurs zero \sis{}, they are $K$-hop indistinguishable.
On the other hand, \sis{} values of true counterparts should be zero for two isomorphic graphs.
In practice, the \sis{} between a true corresponding pairs may only close to 0 due to the structural noise.

Based on \sis{}, we propose a \emph{Structural Inconsistency reducing Graph Matching Algorithm} (SIGMA),
which improves the alignment scores of node pairs that have low \sis{} values in each iteration.
Under suitable assumptions, \method{} can reduce \sis{} values of true counterparts.
As \sis{} considers the multi-hop topological information, \method{} can avoid the misguidance of one-hop structural indistinguishability and find graph matching of higher accuracy.
Besides, we also show that our method increases a soft version of MNC values, which is positively related to MNC and should also be high/low for accurate/inaccurate matching.
Furthermore, we demonstrate that \method{} can be derived by using a mirror descent method to solve the Gromov-Wasserstein distance with a novel $K$-hop-structure-based  matching costs.
As a result, \method{} can be regarded as the first OT-based method that are guaranteed to increase MNC in each iteration in the graph matching literature.

Extensive experimental results demonstrate that the proposed method outperforms state-of-the-art methods.
The rest of the paper is organized as follows.
In Sec. \ref{sec:preliminary}, we review the background and discuss the relationship between \sis{} and matching accuracy.
Based on \sis{}, a new graph matching method \method{} is proposed in Sec. \ref{sec:method}.
We give theoretical analysis in Sec. \ref{sec:theory}.
Due to the limited space, all proof are given in the appendix.
Experimental results are demonstrated in Sec. \ref{sec:experiment} and the appendix.


\section{Preliminary}\label{sec:preliminary}
In this section, we give basic preliminary to facilitate the later discussion and review the main background.

Here, we first give the notations used in the main text. 
We use bold lowercase symbols, bold uppercase letters, and uppercase calligraphic fonts to denote vectors (e.g. $\xB$), matrices (e.g. $\AB$), and sets (e.g. $\AM$), respectively.
$(\cdot)^\top$ is the transpose of a vector or a matrix.
$\AB[i,:]$ and $\AB[:,j]$ are the $i$-th row and the $j$-th column of matrix $\AB$ respectively.
For two matrices $\AB$ and $\BB$ that are of the same size, $\langle \AB,\BB\rangle=\trace(\AB^\top\BB)$ is the Frobenius dot-product.
We denote the cardinality of set $\AM$ by $|\AM|$. 


\subsection{Graph Matching}
We consider undirected graphs without self-loops $\GM=(\VM,\EM,\WB)$, where $\VM$ is the set of vertices, $\EM$ is the set of edges, and  $\WB=[W_{ij}]\in\RBB^{|\VM|\times|\VM|}$ is a (binary or weighted) adjacency matrix.
If there is an edge $(i, j)$ connecting vertices $i$ and $j$, i.e., $(i,j)\in\EM$, the entry $W_{ij}$ represents the weight of the edge;
Otherwise, $W_{ij}=0$.
The degree of each vertex $i$, written as $d_i$, is defined as the sum of the weights of all the edges incident to it, i.e., $d_i=\sum_{j=1}^{|\VM|} W_{ij}$.
The degree matrix $\DB=[D_{ij}]$ has diagonal elements equal to the degrees $D_{ii}=d_i$, and zeros elsewhere.
The Laplacian matrix of $\GM$ is defined as $\LB=\DB-\WB$.
The $k$-hop neighborhood of node $i$, denoted by $\NM_k(i)$, is the subgraph of $\GM$ induced by all nodes that are $k$ or fewer hops away from node $i$.

Given the source graph $\GM^\src=(\VM^\src,\EM^\src,\WB^\src)$ and the target graph $\GM^\tar=(\VM^\tar,\EM^\tar,\WB^\tar)$,
assuming $|\VM^\src|\le|\VM^\tar|$ without loss of generality
graph matching aims to find an injective mapping $\pi:\VM^\src\to\VM^\tar$ that matches nodes in $\GM^\src$ to nodes in $\GM^\tar$.
In this paper, we focus on equal-size graph matching, i.e., $V=|\VM^\src|=|\VM^\tar|$.
In such case, $\pi$ is restricted to be a bijective.
For graphs with different number of nodes, one can add dummy isolated nodes into the smaller graph and make them equal-sized \citep{gold1996graduated,zaslavskiy2008path,jiang2017graph}.
$\pi$ is also commonly represented as a matching matrix $\TB\in\Big\{\TB\in\{0,1\}^{|\VM^\src|\times |\VM^\tar|}|\TB\oneB=\oneB,\TB^\top\oneB=\oneB\Big\}$ (eg. \citep{fey2019deep,heimann2021refining}),
where
\begin{equation}\label{eq:matching matrix}
	T_{ii'}=\begin{cases}
		1,& \text{ if } i \in\VM^\src \text{ is matched to } i'\in\VM^\tar,\\
		0,& \text{ else.}
	\end{cases}
\end{equation}

\subsection{Optimal Transport}
Given two discrete measures $\alpha=\sum_{i=1}^{m}p_i\delta_{\xB_i}(\xB)$ and $\beta=\sum_{j=1}^{n}q_j\delta_{\yB_j}(\yB)$,
Optimal transport (OT) addresses the problem of optimally transporting $\pB=[p_i]\in\RBB_+^m$ toward $\qB=[q_j]\in\RBB_+^n$ \citep{villani2008optimal}.
The $p$-Wasserstein distance between $\pB$ and $\qB$ is defined as
\begin{equation*}
	W_p^p(\pB,\qB)=\min_{\TB\in\Pi(\pB,\qB)}\langle \KB^p,\TB \rangle,
\end{equation*}
where $K_{ij}$ is the distance between $\xB_i$ and $\yB_j$, $\KB^p=[K_{ij}^p]\in\RBB_+^{m\times n}$ and the feasible domain of \emph{transport plan} $\TB$ is given by the set of \emph{coupling measures} $\Pi(\pB,\qB)=\{\TB\in\RBB_+^{m\times n}|\TB\oneB_n=\pB,\TB^\top\oneB_m=\qB\}$.

Gromov-Wasserstein (GW) distance is a generalization of Wasserstein distance \citep{memoli2011gromov}.
Let $\XM$ and $\YM$ be two sample spaces.
Endowing the spaces $\XM$ and $\YM$ with metrics (distances) $d_\XM$ and $d_\YM$,
the GW distance is defined as
\begin{equation*}
	GW_p^p(\pB,\qB)=\min_{\TB\in\Pi(\pB,\qB)}\sum_{i,j=1}^{m}\sum_{i',j'=1}^{n}D_{ii'jj'}^pT_{ii'}T_{jj'},
\end{equation*}
where $D_{ii'jj'}=|d_\XM(\xB_i,\xB_j)-d_\YM(\yB_{i'},\yB_{j'})|$  with $\xB_1,\dots,\xB_m\in\XM$ and $\yB_1,\dots,\yB_n\in\YM$.

Optimal transport can be applied to graph matching.
The source graph $\GM^\src$ and the target graph $\GM^\tar$ are modelled as two probability distributions $\muB^\src=[\mu^\src_i]$ and $\muB^\tar=[\mu^\tar_i]$ respectively,
where
\begin{equation*}
	\mu^z_i=\frac{\sum_{j=1}^{|\VM^z|} W_{ij}^z}{\sum_{i=1}^{|\VM^z|}\sum_{j=1}^{|\VM^z|}W_{ij}^z}, \text{ for }z=\src,\tar.
\end{equation*}
A dissimilarity matrix $\BB^\src$ can be assigned to $\GM^\src$ to encode the structural information.
Each entry $B_{ij}^\src$ models the distance between nodes $i$ and $j$ in $\GM^\src$,
such as the edge weight-based distance \citep{xu2019gromov}, the shortest path distance, or the harmonic distance \citep{verma2017hunt}.
$\BB^\tar$ is defined similarly.
Basically, the node correspondence is estimated by calculating the transport plan of the Gromov-Wasserstein distance that is defined as $GW(\muB^\src,\muB^\tar)=\min_{\TB\in\Pi(\muB^\src,\muB^\tar)}\ell(\TB)$, where
\begin{equation}\label{eq:Xu GW}
	\ell(\TB)=\sum_{i=1}^{|\VM^\src|}\sum_{j=1}^{|\VM^\src|}\sum_{i'=1}^{|\VM^\tar|}\sum_{j'=1}^{|\VM^\tar|}(B_{ij}^\src-B_{i'j'}^\tar)^2T_{ii'}T_{jj'}.
\end{equation}
Node $i\in\VM^\src$ is matched to $\hat{i}\in\VM^\tar$,
where $\hat{i}=\argmax_{i'}T^*_{ii'}$ and $\TB^*$ attains the minimum of Eq. (\ref{eq:Xu GW}).

Note that to better approximate matching matrices (\ref{eq:matching matrix}),
one can relax the feasible domain from $\Big\{\TB\in\{0,1\}^{V\times V}|\TB\oneB=\oneB,\TB^\top\oneB=\oneB\Big\}$ to $\Pi(\oneB,\oneB)=\Big\{\TB\in[0,1]^{V\times V}|\TB\oneB=\oneB,\TB^\top\oneB=\oneB\Big\}$,
instead of $\TB\in\Pi(\muB^\src,\muB^\tar)$.
The matching matrix can then be interpreted as a transport plan from $\oneB$ to $\oneB$.


\subsection{Heat Diffusion Wavelet}

For each graph $z=\src,\tar$,
the heat diffusion on a graph is the solution to the discrete heat equation,
$\vB(t)=\exp(-t\LB^z)\vB(0)$ where $\vB(t)\in\RBB^{|\VM^z|}$ represent the heat of each vertex at time $t$ and $\exp(-t\LB^z)=\sum_{k=0}^{\infty}\frac{(-t)^k}{k!}(\LB^z)^k$ \citep{kondor2002diffusion}.
$\exp(-t\LB)=\sum_{k=0}^{\infty}\frac{(-t)^k}{k!}\LB^k$ is a $|\VM|\times|\VM|$ matrix-valued fuction of $t$, known as the Laplacian exponential diffusion kernel \citep{hammond2013graph}.
As $t$ approaches zero,
high-order terms vanish,
meaning that the kernel depicts \emph{local connectivity} \citep{tsitsulin2018netlsd,donnat2018learning}.
The $K$-th order approximation of $\exp(-t\LB^z)$ is
\begin{equation}\label{eq:heat diffusion wavelet}
	\PsiB=\sum_{k=0}^{K}\frac{(-t)^k}{k!}(\LB^z)^k.
\end{equation}
The $i$-th row, $\PsiB[i,:]$, is the $K$-th order HDW for node $i$ \citep{donnat2018learning} and measures the network connectivity between node $i$ and each node $j$.
$\BB^z=[B_{ij}^z]$ that characterizes node dissimilarities within each graph is defined as
\begin{equation}
	B_{ij}^z=\begin{cases}
		&\bar{\Psi}-\Psi_{ij}^z,\text{ if }i\neq j,\\
		&0, \text{ else,}
	\end{cases}
\end{equation}
where $\Psi_{ij}^z$ is defined as Eq. (\ref{eq:heat diffusion wavelet}), $z=\src,\tar$ and constant $\bar{\Psi}>\max_{i,j,z}\Psi_{ij}^z$.

\section{\SI}\label{sec:si}
Based on the local neighborhoods of node $i$ and node $i'$,
they can be embedded into a $V$-dimensional space,
which yields embedding vectors $\zB_{i}$ and $\zB_{i'}$ respectively.
The $j$-th dimension of $\zB_{i}$ (resp. $j'$-th dimension of $\zB_{i'}$) models the distance between $i$ and $j$ (resp. $i'$ and $j'$).
In this paper,
we instantiate the embeddings based on HDW, i.e., $\zB_{i}=\BB[i:]$ and $\zB_{i'}=\BB[i':]$

Given transport plan $\TB\in\Pi(\oneB,\oneB)$,
we model the \si{} (\sis{}) of a pair of nodes $(i, i')\in\VM^\src\times\VM^\tar$ as a \emph{transport distance}
\begin{equation}
	\masis(i,i';\TB)=\sum_{j=1}^{V}\sum_{j'=1}^{V}T_{jj'}(B_{ij}^\src-B_{i'j'}^\tar)^2.
\end{equation}
\begin{wrapfigure}{r}{0.5\textwidth}
	\begin{subfigure}[b]{0.295\textwidth}
		\includegraphics[width=\textwidth]{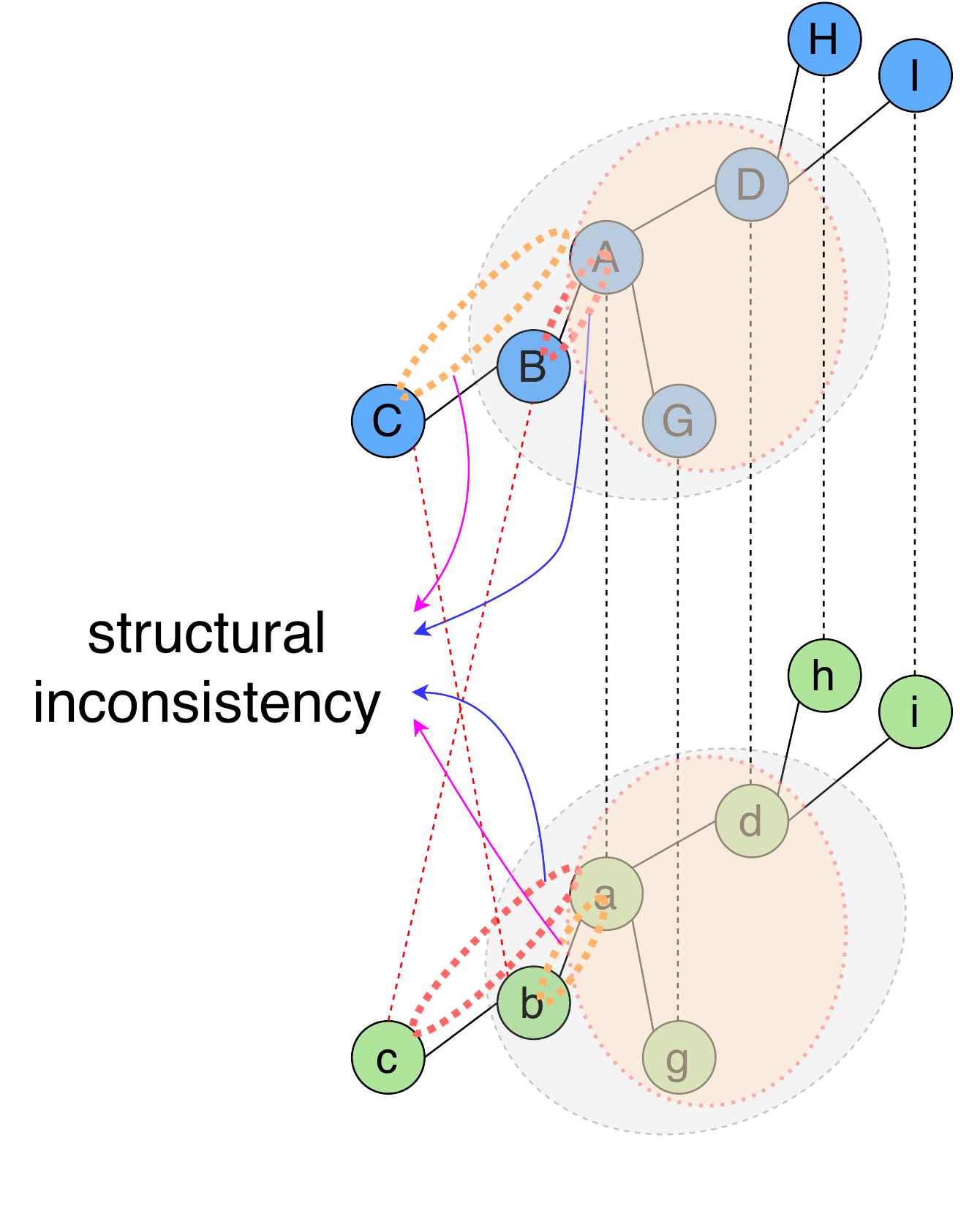}
		\caption{The \si{} of node pair $(A,a)$.}\label{fig:violation part}
	\end{subfigure}
	~
	\begin{subfigure}[b]{0.185\textwidth}
		\includegraphics[width=\textwidth]{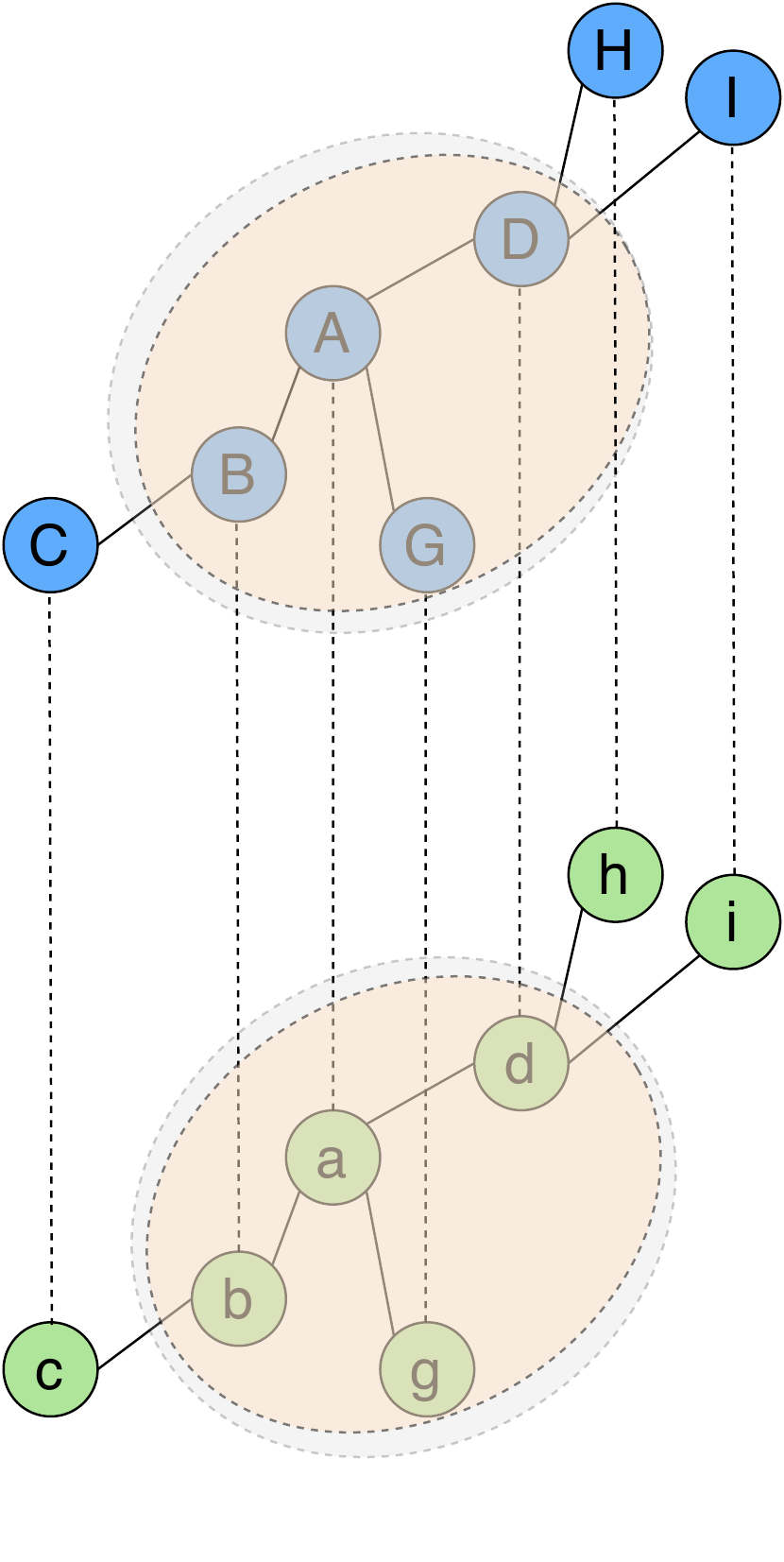}
		\caption{Maximal structural consistency.}\label{fig:conservation part}
	\end{subfigure}
	\caption{The inconsistency of one-hop neighborhood for node pair $(A,a)$. The gray ovals indicate the one-hop neighbors of $A$ and $a$ respectively. The dashed lines denote the matching of $A$ and its one-hop neighbors, and mismatches exist  in Figure \ref{fig:violation part} and thus structural inconsistency occurs.}\label{fig:topology violation}
\end{wrapfigure}
We illustrate $\masis(A,a)$ when $K=1$ in Figure \ref{fig:topology violation}.
Here the transport plans are depicted by the thin dashed lines and we omit them in \sis{}.
In Figure \ref{fig:violation part}, node $B$ (resp. $C$) is wrongly matched to node $c$ (resp. $b$).
Nodes $A$ and $B$ ($a$ and $b$) are connected by an edge while no edge connects $a$ and $c$($A$ and $C$),
which is depicted by the two red(yellow) circles.
Therefore, structural inconsistency incurs.
In Figure \ref{fig:conservation part}, node $A$ and its one-hop neighbors are correctly matched,
which leads to $\masis(A,a)=0$. 

One can derive a \sis{} matrix.
The \sis{} for all node pairs $(i, i')$ given the matching matrix $\TB$, can be written as a matrix $\SB=[S_{ii'}]$ such that $\masis(i,i';\TB)=S_{ii'}$ as
\begin{equation*}
	\SB(\TB)=-2\BB^\src\TB\BB^\tar+h(\BB^\src)\oneB\otimes\oneB+\oneB\otimes h(\BB^\tar)\oneB,
\end{equation*}
where $h$ denotes the element-wise square and $\otimes$ is the outer product of two vectors.
We further define the structural inconsistency of true counterparts (CSI) as
\begin{equation}\label{eq:csi}
	\macsi(i;\TB)=\masis(i,\pi^*(i);\TB),
\end{equation}
where $\pi^*$ is the true matching.

\sis{} is closely related to alignment accuracy.
On the one hand, if node pair $(i,\hat{i})$ have zero \sis{} values, $\hat{i}$ either is the true counterpart $\pi^*(i)$ or have the same local topology as $\pi^*(i)$,
which is formalized in the following theorem.
\begin{theorem}\label{thm:indistinguishable}
	Assume $\GM^\src$ and $\GM^\tar$ are isomorphic and there exists $\TB$ that yields $\masis(i,\hat{i};\TB)=0$.
	Then, if $\hat{i}\neq\pi^*(i)$, $\NM_k^\tar(\hat{i})$ and $\NM_k^\tar\big(\pi^*(i)\big)$ are isomorphic for all $k\le K$.
\end{theorem}
An example of Theorem \ref{thm:indistinguishable} is matching two star graphs, each consisting of one center node connected to $n-1$ peripheral nodes \citep{heimann2021refining}.
Matching peripheral nodes to each other in any order leads to zero \sis{}'s,
whatever the true correspondence of the peripheral nodes is.
This inspires us to use larger neighborhoods to improve the performance, since $\NM_{k+1}^\tar(\hat{i})$ and $\NM_{k+1}^\tar\big(\pi^*(i)\big)$ may not be isomorphic when $\NM_k^\tar(\hat{i})$ and $\NM_k^\tar\big(\pi^*(i)\big)$ are.

On the other hand, if $\GM^\src$ is isomorphic to $\GM^\tar$ up to a few noisy or missing edges (the vetex sets remain the same),
$\macsi\big(i;\TB^*\big)$ values are low where $\TB^*=[T_{ii'}^*]$ is matrix form of true matching.
This is stated in the following theorem.
\begin{theorem}\label{thm:principle}
	Let $\DeltaB_k\in\RBB^{V\times V}$ denote the perturbation to the edge structure, which is given by $\DeltaB_{k}=(\LB^\src)^k-(\tilde{\LB}^\tar)^k$ where $\tilde{\LB}^\tar=\TB^*\LB^\tar\TB^{*\top}$ is the Laplacian matrix of the registered target graph.
	Assume that for all $i$,
	the perturbation to the local neighborhood is bounded as $\epsilon_k=\max_i\sum_{j\in\NM_k^\src(i),j'\in\NM_k^\tar(\pi^*(i))}|\Delta_{k,jj'}|^2$
	Then $\macsi\big(i;\TB^*\big)$ can be bounded from above as follows:
	\begin{equation*}
		\macsi\big(i;\TB^*\big)\le K\sum_{k=1}^{K}\Big(\frac{(-t)^k}{k!}\Big)^2\epsilon_k,
	\end{equation*}
	where $t$ is the propagation time.
\end{theorem}

Note that the propagation time $t$ is often set to a small value in order to capture the local topology  \citep{tsitsulin2018netlsd,donnat2018learning}. 
Thus, for a fixed $K$, when the perturbation $\epsilon_k$ is  small,  $\macsi (i; \TB^* )$ has a small upper bound and even is close enough to 0.
It is reasonable to conjecture that keeping decreasing CSI will lead to an optimal result, which is embodied in our method design.





\section{Methodology}\label{sec:method}


We update the matching matrix based on the \sis{} matrix.
The high level intuition is that each iteration decreases alignment scores for node pairs that have relatively high \sis{} values.
The recurrence of \method{} reads
\begin{equation}\label{eq:iteration}
	\TB^{(\tau+1)}=\proj_{\CM}\TB^{(\tau)}\odot\exp\big(-\eta^{(\tau)}\SB(\TB^{(\tau)})\big),
\end{equation}
where $\tau=0,1,\dots$, $\odot$ is the element-wise multiplication and the exponentiation is also element-wise.
The projection can be solved via Sinkhorn-Knopp algorithm with linear convergence rate \citep{sinkhorn1967concerning}.

Actually, iteration (\ref{eq:iteration}) is the mirror descent update for calculating the GW distance.
To solve problem
\begin{equation}\label{eq:real objective}
	\min_{\TB\in\Pi(\oneB, \oneB)}f(\TB):=\sum_{i=1}^{|\VM^\src|}\sum_{j=1}^{|\VM^\src|}\sum_{i'=1}^{|\VM^\tar|}\sum_{j'=1}^{|\VM^\tar|}(B_{ij}^\src-B_{i'j'}^\tar)^2T_{ii'}T_{jj'},
\end{equation}
the most common algorithm is gradient descent that iteratively updates $\TB$ as follows
\begin{equation}\label{eq:gd}
	\TB^{(\tau+1)}=\proj_{\Pi(\oneB, \oneB)}\Big\{\TB^{(\tau)}-\eta^{(\tau)}\nabla f(\TB^{(\tau)})\Big\},
\end{equation}
where the gradient reads
\begin{equation}\label{eq:gradient Ours}
	\nabla f(\TB)=h(\BB^\src)\oneB\otimes\muB^\src+\oneB\otimes h(\BB^\tar)\muB^\tar-2\BB^\src\TB\BB^\tar,
\end{equation}
Mirror descent takes into account the geometry of the feasible domain \citep{bubeck2015convex}, and is better-suited to Problem (\ref{eq:real objective}).
Mirror descent is given by the recurrence
\begin{equation}\label{eq:md 0}
	\TB^{(\tau+1)}=\argmin_{\TB\in\Pi(\oneB, \oneB)}\langle \nabla f(\TB^{(\tau)}),\TB\rangle+\frac{1}{\eta^{(\tau)}}D_\psi(\TB,\TB^{(\tau)}),
\end{equation}
where $D_\psi(\TB,\TB^{(\tau)})$ stands for the Bregman divergence.
When one selects the generalized KL-divergence $\KL(\AB|\BB)=\sum_{ij}A_{ij}\log\frac{A_{ij}}{B_{ij}}-A_{ij}+B_{ij}$ as the Bregman divergence,
(\ref{eq:md 0}) becomes iteration (\ref{eq:iteration}).

\begin{algorithm}[ht]
	\caption{\method{} \label{alg:method}}
	\begin{algorithmic}[1]
		\State {\textbf{Input:}} Total rounds $T$, graphs $\GM^\src$ and $\GM^\tar$, initial transport plan $\TB^{(0)}$.
		\State {\textbf{Output:}} Correspondence set $\CM$.
		\State Calculate and store matrices $\BB^\src$ and $\BB^\tar$.
		\For{$\tau=0,\dots,T-1$}
		\State Update $\TB^{(\tau+1)}$ by iteration \ref{eq:iteration}.
		\EndFor
		\State $\tilde{\TB}=\TB^{(T)}$
		\State Initialize correspondence set $\CM=\emptyset$
		\For{$i\in\VM^\src$}
		\State $i'=\arg\max_{i'} \tilde{T}_{ii'}$.
		\State $\CM=\CM\cup\{(i,i')\}$
		\EndFor
	\end{algorithmic}
\end{algorithm}

\method{} is summarized in Algorithm \ref{alg:method}.
The matching cost is based on heat diffusion wavelets.
It updates the matching matrix using (\ref{eq:iteration}).
By choosing the largest index from $\tilde{\TB}[i,:]$ for each $i$, we find the correspondence.

\section{Theoretical Analysis}\label{sec:theory}
In this section, we first prove that iteration (\ref{eq:iteration}) can reduce $\macsi(i;\TB)$ for all $i$ under suitable assumptions.
Then the convergence of \method{} is proved.
We also discuss its relationship to state-of-the-art methods.

With appropriate $\eta^{(\tau)}$, iteration \ref{eq:iteration} can reduce alignment score $T_{ii'}$ if $\masis(i,i';\TB^{(\tau)})$ is high,
which is formalized in the following lemma.
\begin{lemma}\label{lem:mirror descent assumptions}
	If $\frac{1}{T_{ii^*}^{(\tau)}}\exp\Big(\eta^{(\tau)}\macsi(i;\TB^{(\tau)})\Big)\le\exp\Big(\eta^{(\tau)}\masis(i;i';\TB^{(\tau)})\Big)$ where $i^*=\pi^*(i)$ and $i'\neq i^*$, $T_{ii'}^{(\tau+1)} < T_{ii'}^{(\tau)}$.
\end{lemma}

We now state our main theorem.
\begin{theorem}\label{thm:each step}
	If $T_{ii'}^{(\tau+1)} < T_{ii'}^{(\tau)}$ for all $i$ and $i'\neq i^*$,
	then $\macsi(i;\TB^{(\tau+1)})< \macsi(i;\TB^{(\tau)})$.
\end{theorem}
Combining Lemma \ref{lem:mirror descent assumptions} and Theorem \ref{thm:each step}, if we have $\frac{1}{T_{ii^*}^{(\tau)}}\exp\Big(\eta^{(\tau)}\macsi(i;\TB^{(\tau)})\Big)\le\exp\Big(\eta^{(\tau)}\masis(i;i';\TB^{(\tau)})\Big)$ for all $i$ and $i'\neq \hat{i}$, we can reduce $\macsi(i;\TB)$ at iteration $\tau$.

By establishing the relationship between \method{} and optimal transport based methods, we can borrow the convergence result of mirror descent and prove that \method{} converges as follows.
\begin{theorem}\label{thm:md}
	\method{} converges to an $\epsilon$-stationary point with the number of mirror descent iterations $\OM(\frac{1}{\epsilon^2})$.
\end{theorem}
The main steps of the proof follow \citet{zhang2018convergence}.
We further remark that the true matching matrix $\TB^*$ is a fixed point of update (\ref{eq:iteration}).


\paragraph{Computational complexity.}
The complexity for calculaing the heat diffusion wavlets is $\OM(KV^3)$.
Due to the two matrix multiplication steps of the term $\BB^\src\TB\BB^\tar$,
calculating the gradient involves computational complexity $\OM(V^3)$.
Note that the matrix multiplication is highly parallelizable and well-suited to modern computing architectures like GPUs.
$\OM(V^2)$ cost is induced by the matrix-vector multiplication in the projection operation.
Therefore, if we run projected mirror descent for $N$ iterations, each of which involves $T$ matrix-vector multiplications in the projection,
$\OM\big(NTV^2+(N+K)V^3\big)$ computational cost is required in learning the matching matrix.
Such complexity is of the same order as state-of-the-art methods like \citet{xu2019gromov,konar2020graph,scott2021graph,heimann2021refining},
and lower than \citet{neyshabur2013netal,malod2015graal,fan2020spectral}.
Considering the sparsity of edges in \realworld{} graphs,
the complexity can be reduced to $\OM\big(NTV^2+(N+K)VE\big)$ where $E=\max\{|\EM^\src|,|\EM^\tar|\}$.

\subsection{Connection to GWL \citet{xu2019gromov}}

The gradient that arises from Problem (\ref{eq:Xu GW}) \citep{xu2019gromov,xu2019scalable,titouan2019optimal,barbe2020graph} and has the following form:
\begin{equation}\label{eq:gradient Xu}
	\nabla\ell(\TB)=h(\BB^\src)\oneB\otimes\muB^\src+\oneB\otimes h(\BB^\tar)\muB^\tar-2\BB^\src\TB\BB^\tar,
\end{equation}
where $h$ denotes the element-wise square and $\otimes$ is the outer product of two vectors \citep{peyre2016gromov}.
Note that if we set $\mu_i^z=1$ for all $i$ and $z=\src,\tar$, (\ref{eq:gradient Xu}) is identical to (\ref{eq:gradient Ours}).

GWL uses \emph{proximal point method} to solve Problem (\ref{eq:Xu GW}) \citep{xu2019gromov}.
If one single mirror descent step is adopted to solve the sub-problem and the marginals are set as $\mu_i^z=1$,
the proximal point method is identical to our mirror descent method.

\subsection{Connection to RefiNA \citet{heimann2021refining}}
When $K=1$, the pairwise \sis{} has a more intuitive interpretation.
\begin{proposition}\label{prop:K=1}
	When one only uses one-hop neighborhood information (i.e., $K=1$),
	\begin{equation*}
		\begin{aligned}
			\exp\big(-\masis(i,i';\TB)\big)=\frac{\exp\Big(t|\tilde{\NM}_1^\src(i;\pi)\cap\NM_1^\tar(i')|\Big)}{\exp\Big(t|\tilde{\NM}_1^\src(i;\pi)\cup\NM_1^\tar(i')|\Big)},
		\end{aligned}
	\end{equation*}
	where $\TB\in\Big\{\TB\in\{0,1\}^{V\times V}|\TB\oneB=\oneB,\TB^\top\oneB=\oneB\Big\}$, $\tilde{\NM}_k^\src(i;\pi)$ is the set of nodes onto which $\pi$ maps $i$'s neighbors $\tilde{\NM}_k^\src(i;\pi)=\{j'\in\VM^\tar|\exists j\in\NM_k^\src(i) \st \pi(j)=j'\}$, and the bijective is given by $\pi(i)=\argmax_{i'}[T_{ii'}]$.
\end{proposition}

\citeauthor{chen2020cone} \citeyear{chen2020cone} and \citeauthor{heimann2021refining} \citeyear{heimann2021refining} define a similar quantity.
Specifically, the \emph{matched neighborhood consistency} (MNC) of node $i$ in $\GM^\src$ and node $i'$ in $\GM^\tar$ is the Jaccard similarity of the sets $\tilde{\NM}_1^{\src}(i;\pi)$ and $\NM_{1}^\tar(i')$, i.e., $\operatorname{MNC}(i, i';\pi)=\frac{\left|\tilde{\NM}_1^{\src}(i;\pi) \cap \NM_{1}^\tar(i')\right|}{\left|\tilde{\NM}_1^{\src}(i;\pi) \cup \NM_{1}^\tar(i')\right|}$.
$\masis(i,i';\TB)$ and $\operatorname{MNC}(i, i';\pi)$ can be related as follows.
\begin{proposition}\label{prop:relationship with MNC}
	When propagation time $t=1$ and $\tilde{\NM}_1^{\src}(i;\pi) \cap \NM_{1}^\tar(i')$ is not empty,
	\begin{equation*}
		\exp\big(-\masis(i,i';\TB)\big) \le \MNC(i,i';\pi).
	\end{equation*}
\end{proposition}

According to Proposition \ref{prop:relationship with MNC}, Lemma \ref{lem:mirror descent assumptions} and Theorem \ref{thm:each step}, \method{} with $K=1$ and $t=1$ can also be viewed as increasing a soft version of MNC values, which is positively related to MNC and should also be high/low for accurate/inaccurate matching.
Combined with the description in Sec. \ref{sec:method}, \method{} can be regarded as the first OT-based method that are guaranteed to increase MNC in each iteration in the graph matching literature.
In addition, by characterizing the structure of a larger neighborhood via HDW, we alleviate the problem of near structural indistinguishability and improves the matching performance.


%
%
%

\section{Experiments}\label{sec:experiment}
In this section, we compare \method{} and state-of-the-art methods.
Due to the limited space, additional experimental results are provided in the appendix.

\begin{table}[tb]
	\centering
	\caption{\Datasets{} used in our experiments.}
	\label{tab:datasets}
	\begin{tabular}{lrrr}
		\toprule
		\textbf{Name}  & \textbf{Nodes} & \textbf{Edges} & \textbf{Description} \\
		\midrule
		\textbf{PPI Yeast}   & 1,004  & 4,920   & protein-protein interaction network \\
		\textbf{Arxiv}            & 18,772  & 198,110 &  collaboration network    \\
		\textbf{LastFM ASIA} & 7,624 & 27,806 & social network  \\
		\bottomrule
	\end{tabular}
\end{table}

\subsection{Experimental Setup}\label{sec:experimental setup}

\paragraph{Computing infrastructure.}
All codes are implemented in Python 3.6 and the package dependencies are listed in the requirements.txt file in the code submission.
The experiments are conducted on a Ubuntu server with two CPUs (Intel Xeon (R) CPU E5-2680 v4 @ 2.40GHz), 4 Nvidia 2080 Ti graphics cards, and 378 GB of RAM.

\paragraph{Baselines.}
Our baselines are unsupervised methods including
(1) RefiNA \citep{heimann2021refining} that also explicitly considers the neighborhood consistency;
(2) REGAL \citep{heimann2018regal} that obtains the node embeddings based on matrix factorization;
(3) GRAMPA \citep{fan2020spectral} that is proposed to match correlated Erd\H{o}s-R\'{e}nyi graphs and involves computing all eigenvalues and the associated eigenvectors of the adjacency matrices;
(4) MM \citep{konar2020graph} that formulates graph matching as maximizing a monotone supermodular set function subject to matroid intersection constraints;
(5) GWL \citep{xu2019gromov} that matches graphs based on OT;
(6) GDD \citep{scott2021graph} that involves solving a three-level nested optimization problem.
As is shown in their papers, RefiNA and MM are sensitive to the noise and thus take the output of CONE-Align \citep{chen2020cone} as initialization.
For fair comparison, GWL and \method{} also use the output of CONE-Align to initialize the transport plan.
We implement REGAL, GWL, and GDD based on the corresponding open-sourced code.

\paragraph{Evaluation Metric.}
We report node correctness ($\maNC$) that is defined as $\maNC=\frac{|\CM\cap\CM_{\mareal}|}{|\CM_{\mareal}|}$,
where $\CM$ and $\CM_{\mareal}$ are the learned and \groundtruth{} node correspondences respectively.




\subsection{Matching Permutated Networks}

\paragraph{\Datasets{}.}
Following the literature (eg. \citep{heimann2018regal,konar2020graph,heimann2021refining}), we match two randomly permutated graphs,
that is,
the target graph is obtained by permutating the nodes of the original graph (source graph).
\method{} is tested against baselines on standard benchmark \datasets{}, including
PPI Yeast \citep{breitkreutz2007biogrid},
Arxiv \citep{leskovec2007graph},
and LastFM ASIA \citep{feather}.
These \datasets{} are listed in Table \ref{tab:datasets}.
Similar to \citet{heimann2018regal,konar2020graph,heimann2021refining},
we also add
structural noise to the target graph by adding $q|\EM^\src|$ edges.

\begin{figure}[ht]
	\centering
	\includegraphics[width=\textwidth]{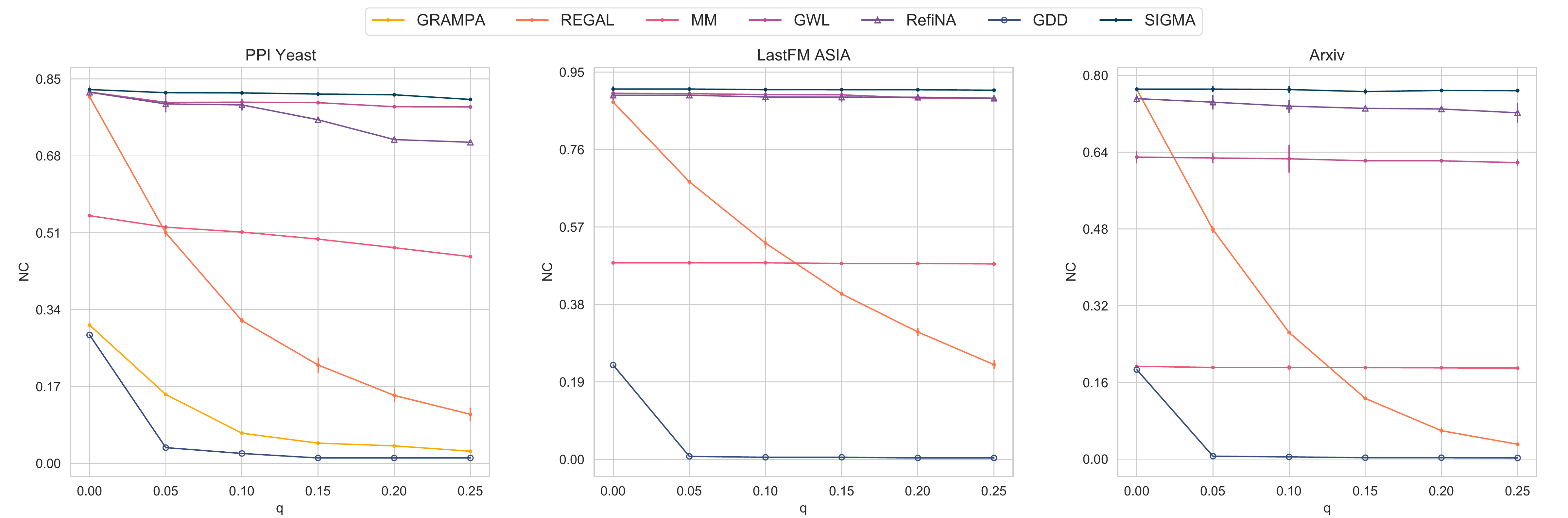}
	\caption{NC of graph matching methods with $q|\EM^\src|$ noisy edges. The errorbars span 2 standard deviation. \method{} outperforms baselines on all \datasets{}.}\label{fig:full main}
\end{figure}

\paragraph{Results.}
We report the performance of \method{} and baselines in Figure \ref{fig:full main},
in which $K$ is set to 3.
The similarity matrix in GRAMPA incurs quadruple computational complexity and takes thousands of hours to obtain on LastFM ASIA and Arxiv.
GRAMPA thus does not scale and its performance is not reported on these two \datasets{}.
OT-based methods and RefiNA take into account the local topology and have good performance on all \datasets{}.
\method{} outperforms state-of-the-art methods in all cases.
The improvement becomes more significant with the noise increasing.

\paragraph{Runtime.}
\begin{table}[tb]
	\centering
	\caption{Average $\pm$ stdev runtime in sec of alignment methods from 5 trials.
		The two fastest methods per dataset are in bold.
		\method{} is faster than its closest competitors in accuracy (Figure \ref{fig:full main}).
	}
	\label{tab:full runtime}
	\begin{tabular}{lrrr}
		\toprule
		\textbf{Dataset}  & \textbf{PPI Yeast} & \textbf{LastFM ASIA} & \textbf{Arxiv} \\
		\midrule
		\textbf{GRAMPA} & 1985.895 $\pm$ 5.825 & / & / \\
		\textbf{REGAL} & 20.579 $\pm$ 0.844 & \textbf{210.842 $\pm$ 14.046} & \textbf{595.431 $\pm$ \quad 18.588}\\
		\textbf{MM} & 20.339 $\pm$ 0.576 & 430.333 $\pm$ 17.871 & 5424.995 $\pm$ \quad 29.580\\
		\textbf{GWL} & 110.682 $\pm$ 0.712 & 967.241 $\pm$ 13.372 & 35682.345 $\pm$ 1024.663\\
		\textbf{RefiNA} & \textbf{17.887 $\pm$ 0.209} & 344.009 $\pm$\,\,\, 8.475 & 3115.087	$\pm$ \quad 74.360 \\
		\textbf{GDD} & 54.666 $\pm$ 1.684 & 6732.156 $\pm$	10.883 & 125965.441 $\pm$	1383.409 \\
		\midrule
		
		\textbf{\method{}} & \textbf{9.731 $\pm$ 0.417} & \textbf{131.241 $\pm$\,\,\, 1.249} & \textbf{2011.765 $\pm$ \quad 15.081} \\
		
		\bottomrule
	\end{tabular}
\end{table}

In Table \ref{tab:full runtime}, we compare the average runtimes of all different methods across noise levels.
Time used to calculate the heat diffusion wavelets in \method{} is included.
The two fastest
methods per dataset are in bold.
 \method{} is faster
than its closest competitors in accuracy (Figure \ref{fig:full main}).

%
%

\section{Related Work}
\paragraph{Matching attributed graphs.}
Deep learning-based methods take as input the node/edge attributes of the graphs and learn node embeddings suitable for graph matching \citep{zanfir2018deep,fey2019deep,wang2019learning,sarlin2020superglue,wang2020combinatorial,yan2020learning}.
These supervised methods, however, require a large amount of anchor links that are often not available in practice.
Optimal transport-based methods exploit structural information and match graphs in an unsupervised manner \citep{maretic2019got,xu2019gromov,chowdhury2019gromov}.
To further incorporate node attributes,
Fused Gromov-Wasserstein distance-based methods are proposed to exploit both node attributes and structure information \citep{titouan2019optimal,barbe2020graph}.
Node or edge attributes can be incorporated into the design of \sis{}.

\paragraph{Graph distance.}
Comparison among graphs is ubiquitous in analyzing graph-structured data.
Spectral distances usually do not take into account all the structural information,
focusing only on the
Laplacian matrix eigenvectors and ignoring a large portion of the structure encoded in eigenvectors \citep{jovanovic2012spectral,gera2018identifying}.
The cut distance \citep{lovasz2012large} and the graph edit distance \citep{bougleux2017graph,raveaux2021unification} require solving difficult discrete optimization problems.
Another family of graph distances that is closely related to this paper is the graph diffusion distance \citep{hammond2013graph,tsitsulin2018netlsd,scott2021graph}.
\citeauthor{hammond2013graph} \citeyear{hammond2013graph} assume the nodes of the two graphs are already matched and calculate the graph diffusion distance.
\citeauthor{scott2021graph} \citeyear{scott2021graph} extend \citet{hammond2013graph}'s work and address the graph matching problem.
Their method involves solving a three-level optimization problem and estimates the node correspondence in the innermost problem by matching eigenvalues of the heat diffusion kernel matrix.
Note that our Eq. (\ref{eq:real objective}) also defines a distance between two graphs.

\section{Conclusion}
In this paper, we consider the problem of matching unattributed graphs without anchor links.
A novel quantity, \sis{}, is defined to measure the correspondence quality of a pair of nodes.
We reveal that \sis{} has a close relationship to matching quality.
On the one hand, if the \sis{} value of a pair of nodes is zero,
they are either true counterparts or have the same local topology.
On the other hand,
true matching entails low \sis{} values of matched node pairs.
Based on \sis{}, we propose a graph matching method named \method{} which can reduce low \sis{} values under suitable assumptions.
Empirical results demonstrate the good performance of our method.

%

\bibliographystyle{plainnat}
\bibliography{ref}

%
%
%
%
%



\newpage

\appendix

\begin{center}
	{\Large \textbf{Appendix}}
\end{center}

This supplementary document contains the technical proofs and some additional numerical results.
Sec. \ref{sec:proof} gives missing proofs in the main paper.
Additional empirical results are demonstrated
in Sec. \ref{sec:add_experiment}.

\section{Technical Proofs}\label{sec:proof}

\subsection{Missing Proofs of Sec. \ref{sec:si}}

%
%
%
%

\paragraph{Proof of Theorem \ref{thm:indistinguishable}.}

Since graph isomorphism is transitive, it suffices to prove that $\NM_K^\src(i)$ is isomorphic to $\NM_K^\tar(\hat{i})$, which we prove by contradiction.
Denote the matching induced by $\TB$ as $\pi:\VM^\src\to\VM^\tar$, i.e., $\pi(i)=\argmax_{i'}T_{ii'}$.
Suppose $\NM_K^\src(i)$ is not isomorphic to $\NM_K^\tar(\hat{i})$.
Then there exists neighboring nodes $a,b\in\NM_K^\src(i)$ where either $\pi(a)$ or $\pi(b)$ is not in $\NM_K^\tar(\hat{i})$, or $\pi(a)$ and $\pi(b)$ do not share an edge.

In case 1, without loss of generality, $\pi(b)\notin\NM_K^\tar(\hat{i})$.
Then $\masis(i,\hat{i};\TB)\ge T_{b\pi(b)}(B_{ib}^\src-B_{\hat{i}\pi(b)}^\tar)^2>0$: a contradiction.

In case 2, since $\pi(b)$ is the counterpart of a neighbor of $a$, it must also be a neighbor of the counterpart of $a$ which is a contradition of the assumption that $\pi(a)$ and $\pi(b)$ do not share an edge,
or else $\masis\big(a,\pi(a)\big)>0$, another contradiction.

Therefore, we conclude that $\NM_K^\src(i)$ and $\NM_K^\tar(\hat{i})$ are isomorphic.

\begin{flushright}
	$\blacksquare$
\end{flushright}

\paragraph{Proof of Theorem \ref{thm:principle}.}


For \groundtruth{} node pairs $(i,i^*)$ and \groundtruth{} matching matrix, denoting $a_k=\frac{(-t)^k}{k!}$, we have
\begin{equation*}
	\begin{aligned}
		\masis(i,i^*;\TB^*)&=\sum_{j=1}^{|\VM^\tar|}(B_{ij}^\src-B_{i^*j^*}^\tar)^2\\
		&\overset{\textcircled{1}}{=}\sum_{j=1}^{|\VM^\tar|}\Big(\sum_{k=1}^{K}a_k[(\LB^\src)^k]_{ij}-a_k[\TB^*(\LB^\tar)^k\TB^{*\top}]_{ij}\Big)^2\\
		&\overset{\textcircled{2}}{\le}\sum_{j=1}^{|\VM^\tar|}K\sum_{k=1}^{K}(a_k)^2\Big([(\LB^\src)^k]_{ij}-[\TB^*(\LB^\tar)^k\TB^{*\top}]_{ij}\Big)^2\\
		&\overset{\textcircled{3}}{\le}K\sum_{k=1}^{K}(a_k)^2\epsilon_k,
	\end{aligned}
\end{equation*}
where we substitute the definitions of $B_{ij}^\src$ and $B_{i^*j^*}^\tar$ into equality \textcircled{1},
use the Cauchy–Schwarz inequality in inequality \textcircled{2},
and use the definition of $\epsilon_k$ in inequality \textcircled{3}.
\begin{flushright}
	$\blacksquare$
\end{flushright}

\subsection{Missing Proofs of Sec. \ref{sec:theory}}

\paragraph{Proof of Lemma \ref{lem:mirror descent assumptions}.}

The mirror descent recurrence is $\TB^{(\tau+1)}=\proj_{\Pi(\oneB,\oneB)}\Big(\TB^{(\tau)}\odot \GB(\TB^{(\tau)})\Big)$,
where $\GB(\TB^{(\tau)})=\exp\Big(-\eta^{(\tau)}h(\BB^\src)\oneB\otimes\oneB-\eta^{(\tau)}\oneB\otimes h(\BB^\tar)\oneB+2\eta^{(\tau)}\BB^\src\TB^{(\tau)}\BB^\tar\Big)$.
The projection is achieved by Sinkhorn scaling \citep{knight2008sinkhorn},
that is,
\begin{equation*}
	\TB^{(\tau+1)}=\Diag(\uB)\YB\Diag(\vB),
\end{equation*}
where $\YB=\TB^{(\tau)}\odot \GB(\TB^{(\tau)})$, and $\uB,\vB\in\RBB_+^{V}$.

The recurrence implies that,
\begin{equation}\label{eq:beta}
	Y_{ii'}= T_{ii'}^{(\tau)}\exp\big(-\eta^{(\tau)}\masis(i,i';\TB^{(\tau)})\big).
\end{equation}

The feasible domain requires that
\begin{equation*}
	1=\sum_{i'}u_iY_{ii'}v_{i'}, \text{ and }1=\sum_{i}u_iY_{ii'}v_{i'}.
\end{equation*}
Hence,
\begin{equation}\label{eq:alpha}
	v_{i'}=\frac{1}{\sum_{i}u_iY_{ii'}}\le\frac{1}{u_iY_{ii^*}}= \frac{1}{u_iT_{ii^*}^{(\tau)}\exp\Big(-\eta^{(\tau)}\macsi(i;\TB^{(\tau)})\Big)},
\end{equation}
where the first inequality is due to the fact that $u_i\ge0$ and $Y_{ii'}\ge0$.

Combining Eq. (\ref{eq:alpha}), (\ref{eq:beta}), and $\frac{1}{T_{ii^*}^{(\tau)}}\exp\Big(\eta^{(\tau)}\macsi(i;\TB^{(\tau)})\Big)\le\exp\Big(\eta^{(\tau)}\masis(i;i';\TB^{(\tau)})\Big)$,
we have $T_{ii'}^{(\tau+1)}\le T_{ii'}^{(\tau)}$ for $i'\neq i^*$.


\begin{flushright}
	$\blacksquare$
\end{flushright}

\paragraph{Proof of Theorem \ref{thm:each step}.}

By the definition of the \si{},
\begin{equation*}
	\begin{aligned}
		\masis(i,i^*;\TB)&=\sum_{j\in\NM_K^\src(i)}T_{jj^*}(B_{ij}^\src-B_{i^*j^*}^\tar)^2+\sum_{j\notin\NM_K^\src(i)}T_{jj^*}(B_{ij}^\src-B_{i^*j^*}^\tar)^2+\sum_{j=1}^{|\VM^\src|}\sum_{j'\neq j^*}T_{jj'}(B_{ij}^\src-B_{i^*j'}^\tar)^2\\
		&=\sum_{j=1}^{|\VM^\src|}\sum_{j'\neq j^*}T_{jj'}(B_{ij}^\src-B_{i^*j'}^\tar)^2.
	\end{aligned}
\end{equation*}
Where the second equality is due to the fact that $\NM_K^\src(i)$ is isomorphic to $\NM_K^\tar(i^*)$ and $B_{ij}^\src=B_{i^*j^*}^\tar=\bar{\Psi}$ if node $j$ is more than $K$ hops away from node $i$.
If $T_{ii'}^{(\tau+1)}\le T_{ii'}^{(\tau)}$ for all $i$ and $i'\neq i^*$,
\begin{equation*}
	\masis(i,i^*;\TB^{(\tau+1)})=\sum_{j=1}^{|\VM^\src|}\sum_{j'\neq j^*}T_{jj'}^{(\tau+1)}(B_{ij}^\src-B_{i'j^*}^\tar)^2\le \sum_{j=1}^{|\VM^\src|}\sum_{j'\neq j^*}T_{jj'}^{(\tau)}(B_{ij}^\src-B_{i'j^*}^\tar)^2.
\end{equation*}
The right hand side of the inequality is exactly $\masis(i,i^*;\TB^{(\tau)})$.


\begin{flushright}
	$\blacksquare$
\end{flushright}

%
%
%

\paragraph{Proof of Theorem \ref{thm:md}}
We can rewrite the transport plan $\TB$ as a vector variable $\xB\in\RBB_+^{V^2}$ and the Gromov-Wasserstein distance Eq. (\ref{eq:real objective}) as objective $f(\xB)=\xB^\top\AB\xB$, respectively.

It is easy to verify that
\begin{enumerate}
	\item The feasible domain is a closed convex set.
	\item $f(\xB)$ is a weakly convex function.
	\item The norm of the gradient $\|\nabla f(\xB)\|$ for all feasible $\xB$ is bounded.
	\item The optimal objective value, denoted as $f_{\min}$, exists and $f_{\min}>-\infty$.
\end{enumerate}

By Corollary 3.1 of \citet{zhang2018convergence}, \method{} converges to an $\epsilon$-stationary point with the number of mirror descent iterations $\OM(\frac{1}{\epsilon^2})$.

\begin{flushright}
	$\blacksquare$
\end{flushright}

\paragraph{Proof of Proposition \ref{prop:K=1}}
When $K=1$, the \sis{} matrix can be rewritten as
\begin{equation*}
	\SB(\TB)=-t\WB^\src\TB\WB^\tar+t(\WB^\src\otimes\oneB+\oneB\otimes\WB^\tar-\WB^\src\TB\WB^\tar).
\end{equation*}
Note that $\tilde{\NM}_1^\src(i;\pi)\cap\NM_1^\tar(i')=\{j'|W_{ij}^\src T_{jj'}W_{i'j'}\neq 0\}$. For $\TB\in\Big\{\TB\in\{0,1\}^{V\times V}|\TB\oneB=\oneB,\TB^\top\oneB=\oneB\Big\}$, we have
\begin{equation*}
	|\tilde{\NM}_1^\src(i;\pi)\cap\NM_1^\tar(i')|=[\WB^\src\TB\WB^\tar]_{ii'}.
\end{equation*}
Since
\begin{equation*}
	|\tilde{\NM}_1^\src(i;\pi)\cup\NM_1^\tar(i')|=|\tilde{\NM}_1^\src(i;\pi)|+|\NM_1^\tar(i')|-|\tilde{\NM}_1^\src(i;\pi)\cap\NM_1^\tar(i')|,
\end{equation*}
we have
\begin{equation*}
	\begin{aligned}
		-\masis(i,i';\TB)&=t[\WB^\src\TB\WB^\tar]_{ii'}-t[\WB^\src\otimes\oneB+\oneB\otimes\WB^\tar-\WB^\src\TB\WB^\tar]_{ii'}\\
		&=t|\tilde{\NM}_1^\src(i;\pi)\cap\NM_1^\tar(i')|-t|\tilde{\NM}_1^\src(i;\pi)\cup\NM_1^\tar(i')|
	\end{aligned}
\end{equation*}
\begin{flushright}
	$\blacksquare$
\end{flushright}

To prove Proposition \ref{prop:relationship with MNC}, we need the following lemma.

\begin{lemma}\label{lem:inequality}
	For $1\le a\le b$, $\frac{a}{b}\ge\frac{e^a}{e^b}$.
\end{lemma}
\textbf{Proof:}

Consider function $f(x)=x-\log x$.
The derivative $f'(x)=1-\frac{1}{x}\ge 0$, for all $x\ge 1$.
Therefore, $f(x)$ increases monotonically for all $x\ge 1$.

Thus, $b-\log b\ge a - \log a$ for $1\le a\le b$, which can be rearranged to $\frac{a}{b}\ge\frac{e^a}{e^b}$.

\begin{flushright}
	$\blacksquare$
\end{flushright}

We now prove Proposition \ref{prop:relationship with MNC}.

\paragraph{Proof of Proposition \ref{prop:relationship with MNC}.}

%

By Proposition \ref{prop:K=1}, when $K=1$ and $\TB\in\Big\{\TB\in\{0,1\}^{V\times V}|\TB\oneB=\oneB,\TB^\top\oneB=\oneB\Big\}$,
\begin{equation*}
	\exp\big(-\masis(i,i';\TB)\big)=\frac{\exp\Big(t|\tilde{\NM}_1^\src(i;\pi)\cap\NM_1^\tar(i')|\Big)}{\exp\Big(t|\tilde{\NM}_1^\src(i;\pi)\cup\NM_1^\tar(i')|\Big)}.
\end{equation*}
By Lemma \ref{lem:inequality}, when $|\tilde{\NM}_1^\src(i;\pi)\cap\NM_1^\tar(i')|\ge 1$ and $t=1$, we have the result.

\begin{flushright}
	$\blacksquare$
\end{flushright}

\section{Additional Experimental Results}\label{sec:add_experiment}
\subsection{Additional Experiment Details}\label{sec:additional experimental setup}

We first include more experimental details about parameter choice in this subsection.
For experiments in the main paper, we set the propagation time $t=1\times 10^{-3}$ and the number of hops $K=3$.


\paragraph{The influence of $K$.}

\begin{figure}[ht]
	\centering
	\includegraphics[width=\textwidth]{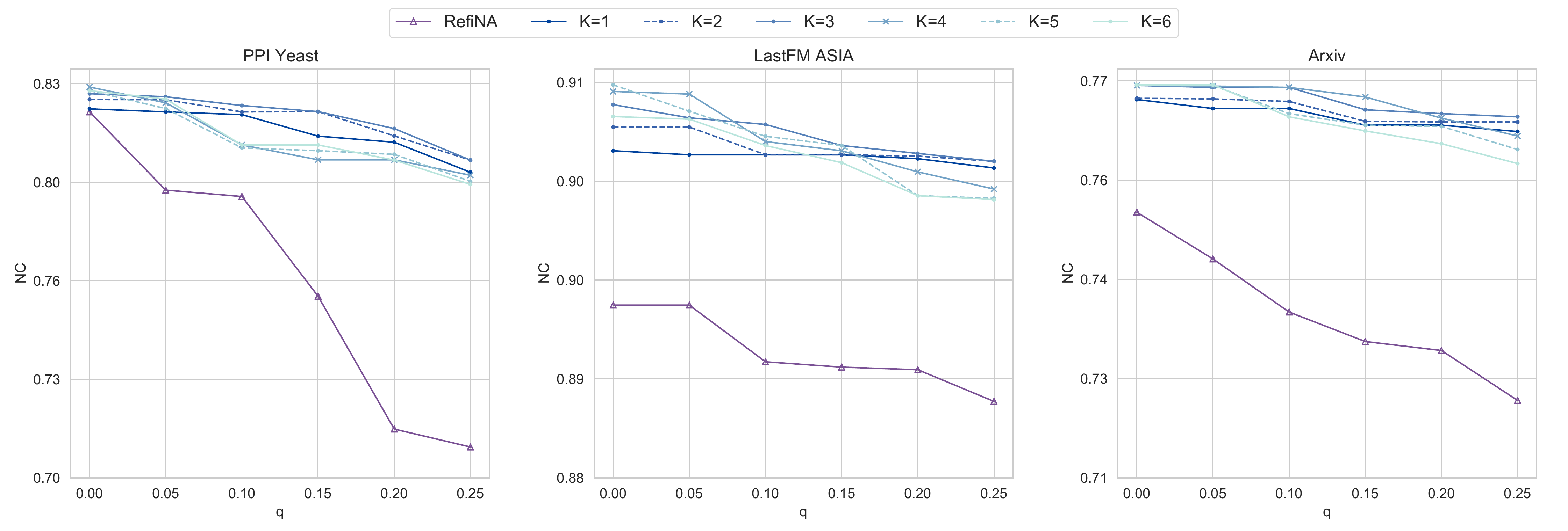}
	\caption{NC of \method{} for different $K$ and $q$. The propagation time $t$ is set as $1\times 10^{-3}$.}\label{fig:n_hop}
\end{figure}

We empirically study the influence of $K$.
On the one hand, increasing $K$ can increase the distinguishability of nodes and thus boost the matching accuracy.
On the other hand, \method{} with a larger $K$ is easier to be affected by structural noise, as is implied by the upper bound in Theorem \ref{thm:principle}.
Figure \ref{fig:n_hop} demonstrates that $K=3$ can achieve a balance.
With the amount of noise increasing, \method{} with a larger $K$ deteriorates faster.
\method{} with $K=1$ outperforms RefiNA, which demonstrates that \method{} is more related to MNC theoretically.

\paragraph{Sensitivity to propagation time $t$.}

\begin{figure}[ht]
	\centering
	\includegraphics[width=\textwidth]{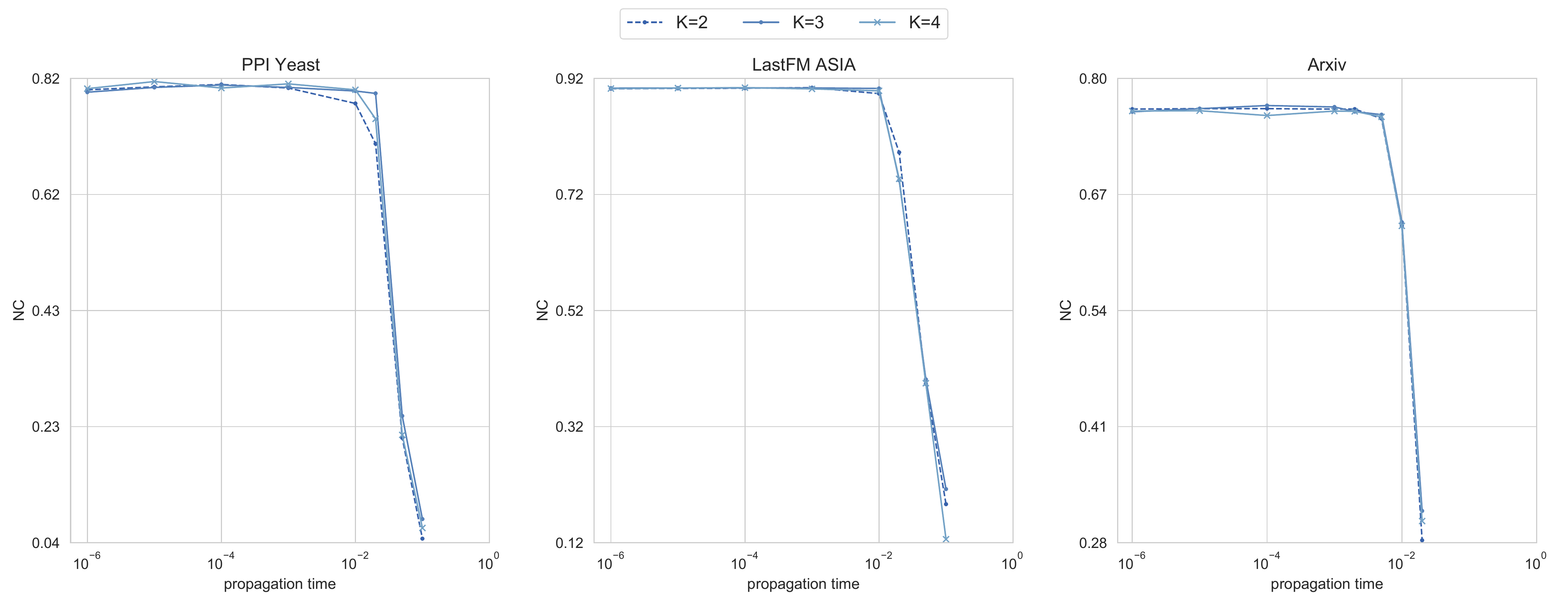}
	\caption{NC of \method{} with varying values of propagation time $t$ when $q=0.25$ for $K=2,3,4$.}\label{fig:prop_time}
\end{figure}

The range of propagation time is determined by line search, which is demonstrated in Figure \ref{fig:prop_time}.
On all \datasets{},
\method{} is not sensitive to the choice of propagation time $t$ and achieves good performance when $1\times 10^{-6}\le t\le 1\times 10^{-3}$.
Note that when $t$ is too small, $\PsiB\simeq\IB-t\LB$ where $\IB$ is the identity matrix and \method{} is reduced to only considering one-hop information.
Therefore, the accuracy slightly drops when $t\le 1\times 10^{-5}$.

\subsection{Matching \Realworld{} Related networks}
%
%



\begin{table}[ht]
	\centering
	\caption{Statistics of Oregon Route Views graphs.}\label{tb:oregon data}
	\begin{tabular}{c | c | c c c c c}
		\hline
		\multirow{2}{*}{\textbf{Dates of RVs}} & \textbf{Source Graph}  & \multicolumn{5}{|c}{\textbf{Target Graphs}}                                          \\
		& Mar. 31\textsuperscript{st} & Mar. 31\textsuperscript{st} & Apr. 7\textsuperscript{th} & Apr. 14\textsuperscript{th} & Apr. 21\textsuperscript{st} & Apr. 28\textsuperscript{th} \\
		\hline
		\textbf{Nodes}                        & 10,670        & 10,670        & 10,729       & 10,790       & 10,859       & 10,886       \\
		\textbf{Edges}                        & 22,002        & 22,002        & 21,999       & 22,469       & 22,747       & 22,493      \\
		\textbf{Diameters} & 9 & 9 & 11 & 9 & 10 & 10  \\
		\hline
	\end{tabular}
\end{table}

We further demonstrate the performance of \method{} by matching \realworld{} correlated graphs, i.e., the structural difference between the two graphs is induced by \emph{real noise} instead of \emph{simulated noise}.
The experiments are conducted on graphs of Autonomous Systems peering information inferred from Oregon Route-Views (RV) \footnote{\url{http://snap.stanford.edu/data/Oregon-1.html}} between March 31\textsuperscript{st}, 2001 and April 28\textsuperscript{th}, 2001.
The University's RV project\footnote{\url{http://www.routeviews.org/routeviews/}} was a tool for Internet operators to obtain real-time border gateway protocol information about the global routing system.
We match RV on March 31\textsuperscript{st} to RVs between March 31\textsuperscript{st} and April 28\textsuperscript{th}. The \dataset{} statistics are listed in the Table \ref{tb:oregon data}.
We still set $t=1\times 10^{-3}$ and $K=3$ for \method{}.
For fair comparison, RefiNA, MM, GWL, and \method{} use the ouput of CONE-Align to initialize the matching matrix.

We report the results on Table \ref{tb:oregon results}.
As RV changes over time, the  performance of all methods decreases except the one on April 28\textsuperscript{th}.
This is possibly due to monthly patterns in RVs.
\method{} outperforms all baselines.
GDD does not take as input the result of CONE-Align and is a deterministic algorithm. The standard deviation is thus 0.
Other 0.000 deviation values are induced by rounding to three decimal places.

\begin{table}[ht]
	\centering
	\caption{NCs on Oregon Route Views graphs across 5 trials.}\label{tb:oregon results}
	\begin{tabular}{c c c c c c}
		\toprule
		$\GM^\tar$ & Mar. 31\textsuperscript{st}     & Apr. 7\textsuperscript{th}      & Apr. 14\textsuperscript{th}     & Apr. 21\textsuperscript{st}     & Apr. 28\textsuperscript{th}      \\
		\midrule
		\textbf{REGAL}         & 0.491 $\pm$ 0.000 & 0.232 $\pm$ 0.001 & 0.189 $\pm$ 0.000 & 0.158 $\pm$ 0.000 & 0.115 $\pm$ 0.000 \\
		\textbf{MM}            & 0.506 $\pm$ 0.004 & 0.314 $\pm$ 0.001 & 0.202 $\pm$ 0.001 & 0.132 $\pm$ 0.000 & 0.173 $\pm$ 0.001  \\
		\textbf{GWL}           & 0.508 $\pm$ 0.002 & 0.362 $\pm$ 0.001 & 0.256 $\pm$ 0.001 & 0.186 $\pm$ 0.000 & 0.229 $\pm$ 0.000  \\
		\textbf{RefiNA}        & 0.259 $\pm$ 0.005 & 0.172 $\pm$ 0.000 & 0.169 $\pm$ 0.000 & 0.110 $\pm$ 0.000 & 0.124 $\pm$ 0.000  \\
		\textbf{GDD}           & 0.341 $\pm$ 0.000 & 0.007 $\pm$ 0.000 & 0.006 $\pm$ 0.000 & 0.007 $\pm$ 0.000 & 0.007 $\pm$ 0.000  \\
		\midrule
		\textbf{\method{}}         & \textbf{0.514 $\pm$ 0.002} & \textbf{0.391 $\pm$ 0.000} & \textbf{0.264 $\pm$ 0.001} & \textbf{0.198 $\pm$ 0.000} & \textbf{0.235 $\pm$ 0.000} \\
		\bottomrule
	\end{tabular}
\end{table}

\end{document}